\documentclass[10pt,twocolumn,letterpaper]{article}

\usepackage{wacv}
\usepackage{booktabs}
\usepackage{multirow}
\usepackage{graphicx}
\usepackage[table]{xcolor}
\usepackage{placeins}   
\usepackage[normalem]{ulem}
\usepackage{cuted}
\usepackage{caption}
\usepackage{fontawesome5}

\usepackage{xcolor}

\newcommand{\sg}{\operatorname{sg}}

\definecolor{IllinoisOrange}{HTML}{FF5F05}

\definecolor{wacvblue}{rgb}{0.21,0.49,0.74}
\usepackage[pagebackref,breaklinks,colorlinks,allcolors=wacvblue]{hyperref}

\def\wacvPaperID{1804}
\def\confName{WACV}
\def\confYear{2027}

\definecolor{illinoisorange}{HTML}{FF5F05}

\begin{document}

\title{From Tokens to Trajectories: Parameter-Efficient Continuous Adaptation of Unified Multimodal Models}

\author{
    Gayatri Deshmukh \quad
    Onkar Susladkar \quad
    Adheesh Juvekar\\
    Halil Kilicoglu\textsuperscript{\ensuremath{\dagger}} \quad
    Ismini Lourentzou\textsuperscript{\ensuremath{\dagger}}\\[4pt]
    {University of Illinois Urbana-Champaign}\\[2pt]
    {\small\ttfamily \{gsd4,onkarks2\}@illinois.edu}\\[2pt]
    {\href{https://plan-lab.github.io/FlowShift/}{\textcolor{illinoisorange}{plan-lab.github.io/FlowShift}}}\\[3pt]
    {\small \textsuperscript{\ensuremath{\dagger}}Corresponding authors}
}

\providecommand{\ours}{{\textit{FlowShift}}}

\providecommand{\oursbold}{\textbf{\textit{FlowShift}}}

\twocolumn[{%
\renewcommand\twocolumn[1][]{#1}%
\maketitle
\begin{center}
\vspace{-0.8cm}
    \centering
\includegraphics[width=0.96\textwidth]{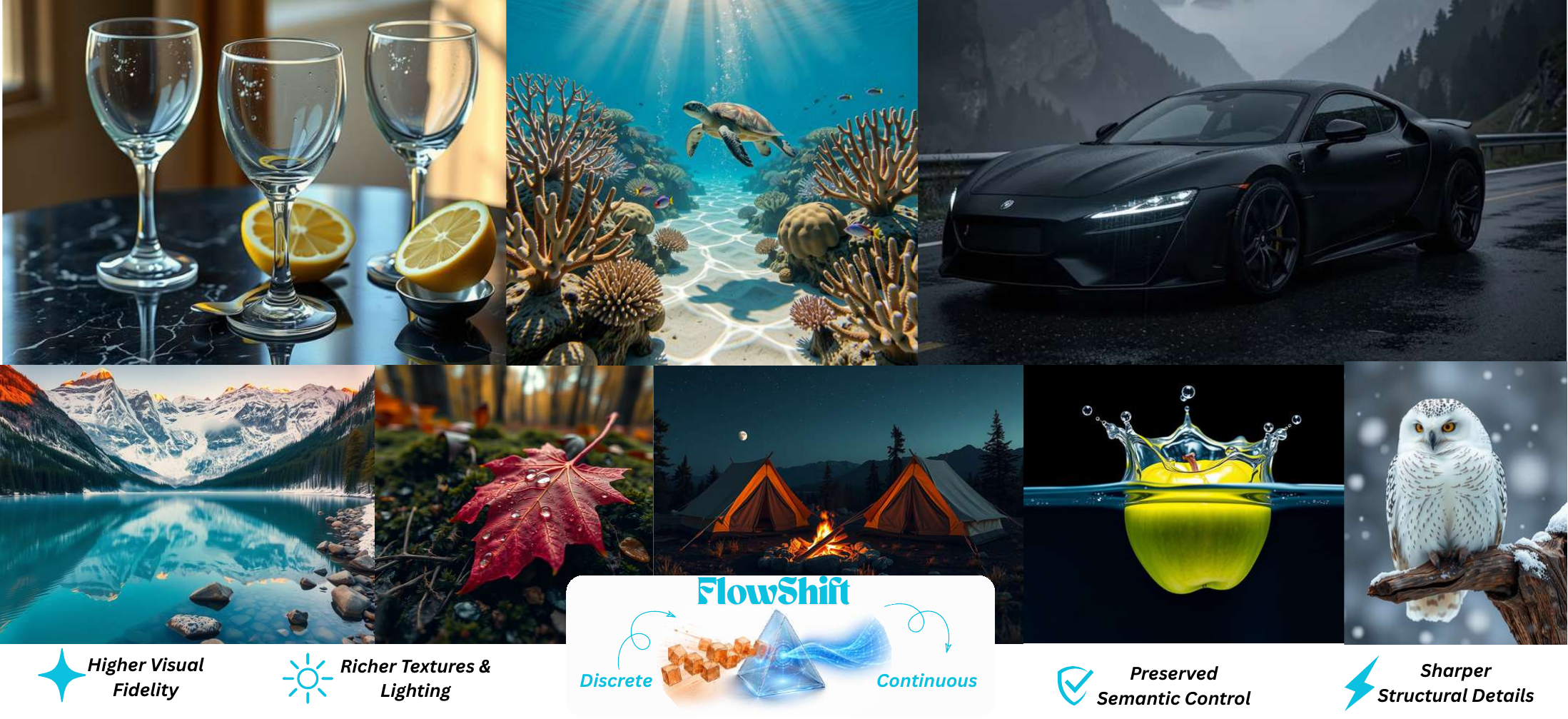}
    \vspace{-0.3cm}
    \captionof{figure}{
    \ours{} converts discrete unified generators to continuous flow, improving fidelity while preserving compositional control.
    }
    \vspace{0.3cm}
    \label{fig:teaser}
\end{center}
}]
\newcommand{\MethodologyFigure}{
\begin{figure*}[t]
    \centering
    \includegraphics[width=0.99\linewidth]{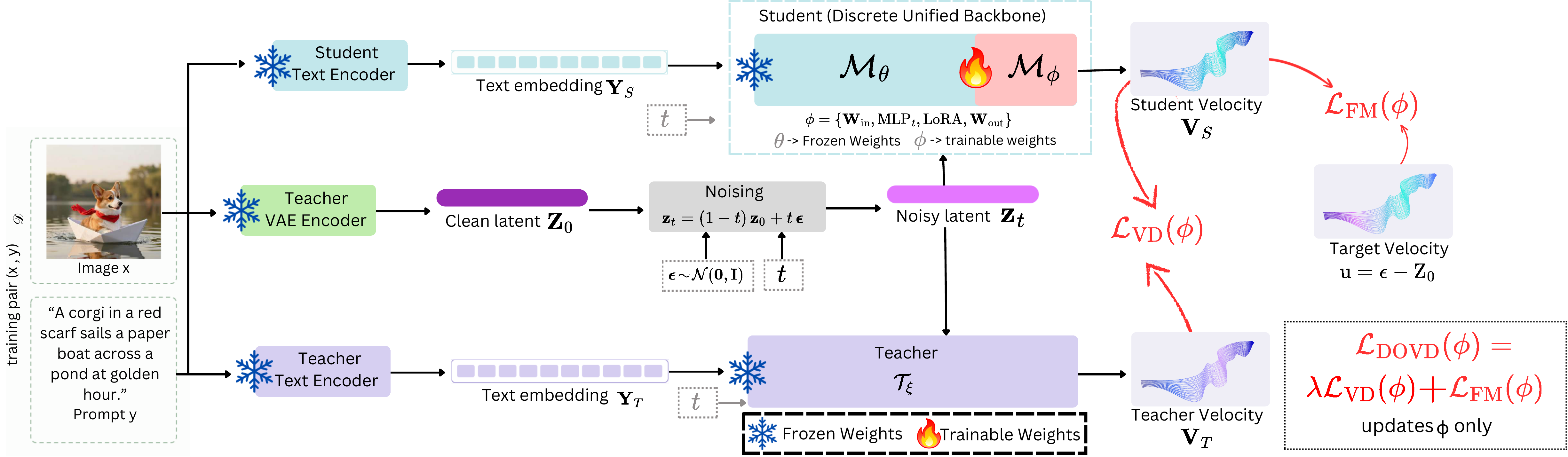}
    \vspace{-.3cm}
    \caption{\small \ours{}: a frozen discrete UMM is adapted to continuous latent flow using lightweight trainable modules and dual-objective velocity distillation from a frozen continuous teacher.}
    \label{fig:MethodologyFigure}
    \vspace{-10pt}
\end{figure*}
}

\newcommand{\UniDFlowFigure}{
\begin{figure*}[!t]
    \centering
    \includegraphics[
        width=0.97\textwidth,
        height=0.99\textheight,
        keepaspectratio
    ]{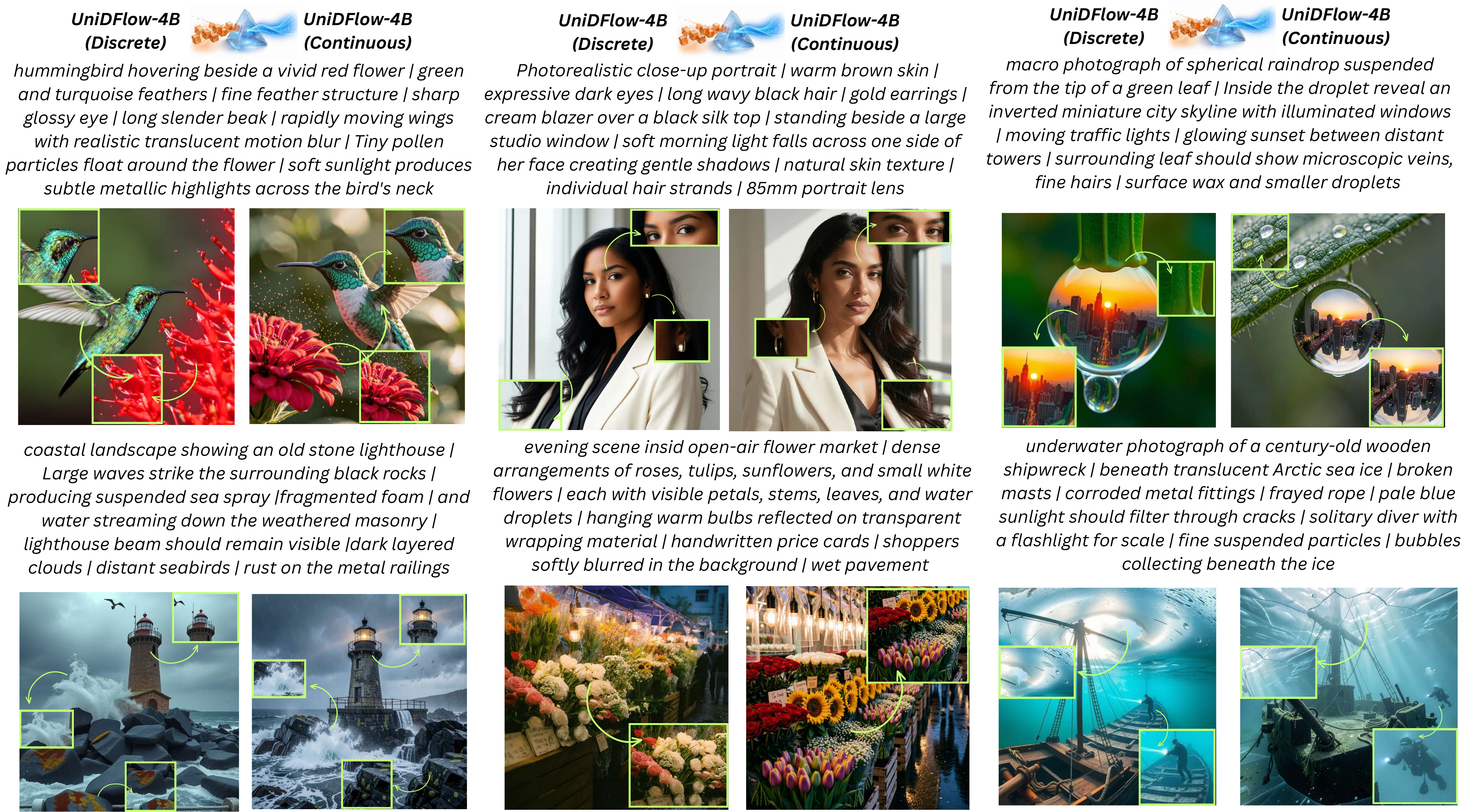}
    \vspace{-9pt}
    \caption{\small Qualitative comparison of discrete and continuous generation with UniDFlow-4B.
    }
    \label{fig:UniDFlowFigure}
    \vspace{-0.2cm}
\end{figure*}
}

\newcommand{\FUDOKIFigure}{
\begin{figure*}[!t]
    \centering
    \includegraphics[
        width=0.97\textwidth,
        height=0.99\textheight,
        keepaspectratio
    ]{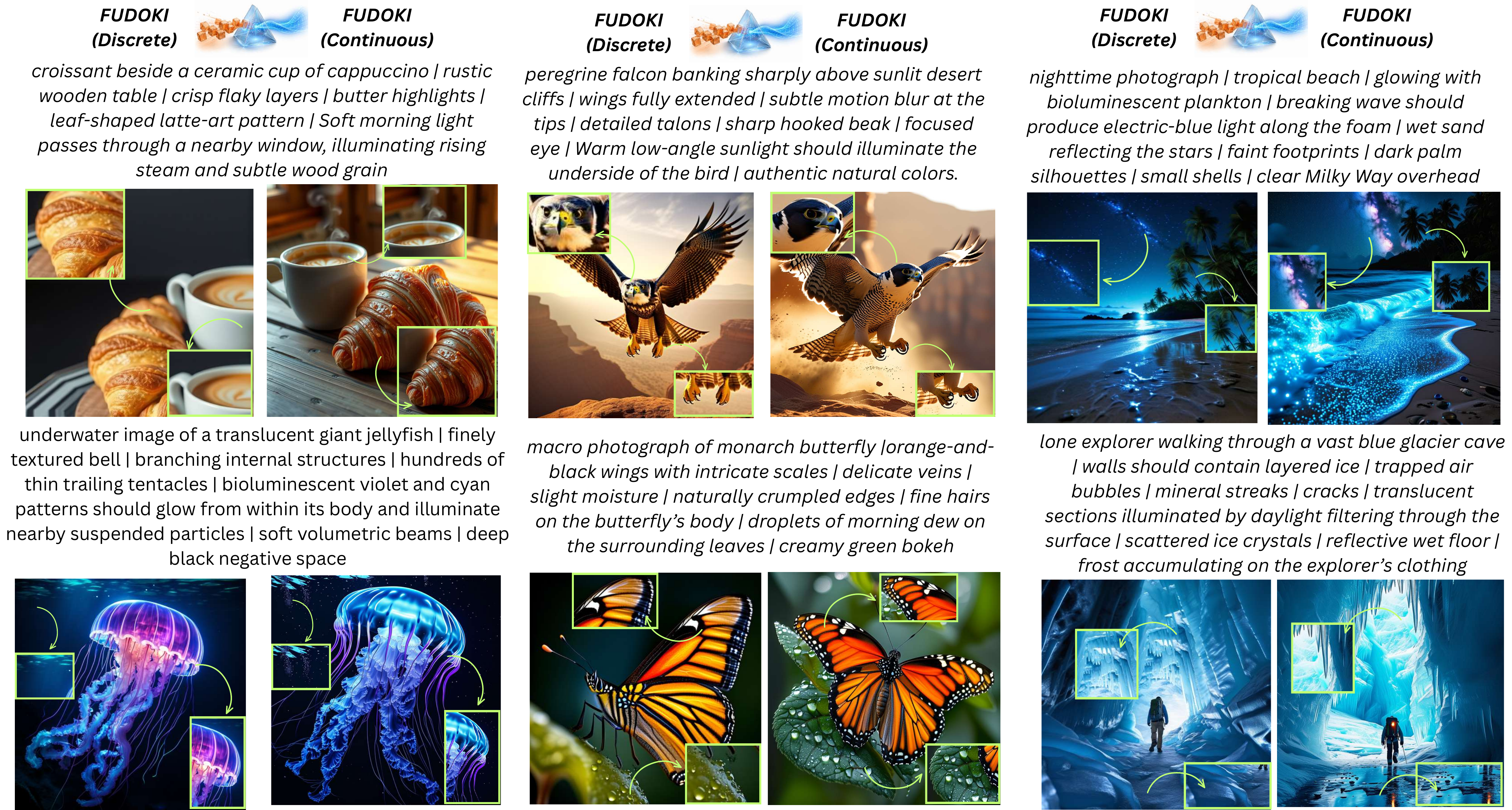}
    \vspace{-9pt}
    \caption{\small Qualitative comparison of discrete and continuous generation of FUDOKI.
    }
    \label{fig:FUDOKIFigure}
        \vspace{-0.2cm}
\end{figure*}
}

\newcommand{\UndResultsFigure}{
\begin{figure*}[!t]
    \centering
    \includegraphics[width=0.97\linewidth]{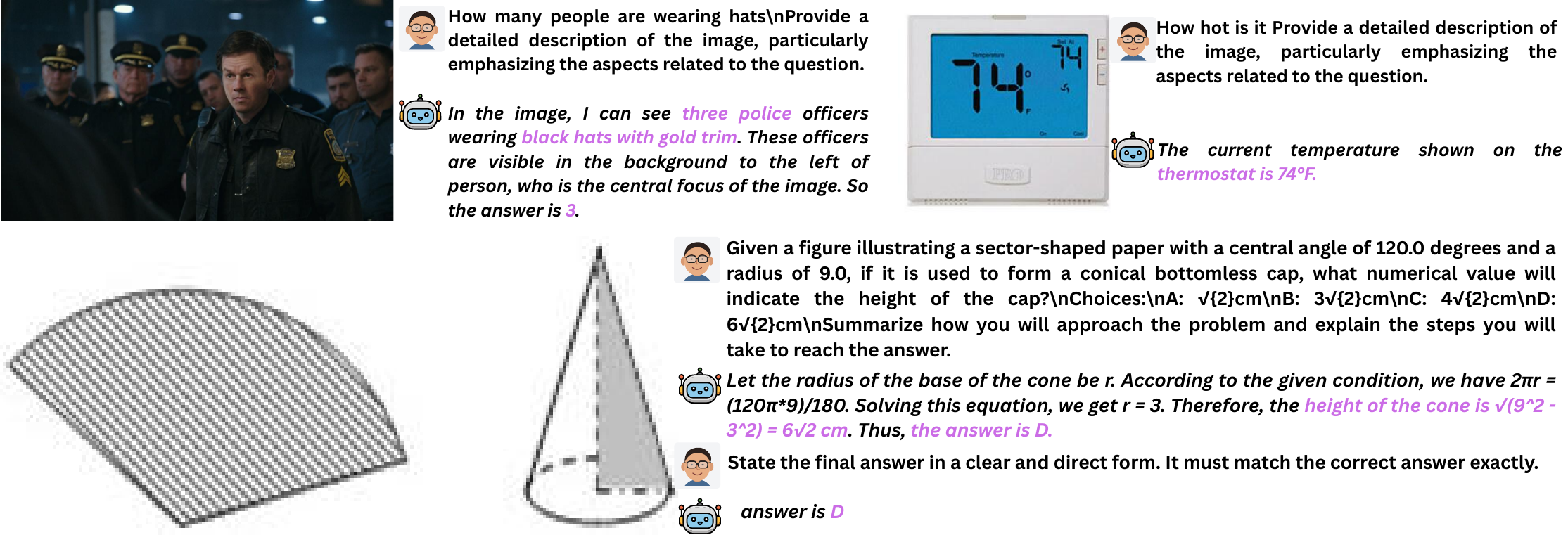}
     \vspace{-.3cm}
    \caption{\small VQA/Image understanding using continuous UniDFlow-4B. The examples demonstrate that \ours{} successfully preserves complex reasoning capabilities despite the transition from a discrete to a continuous visual generation space.}
    \label{fig:UndResults}
    \vspace{-0.3cm}
\end{figure*}
}

\newcommand{\SuppFigure}{
\begin{strip}
    \centering
    \includegraphics[
        width=\textwidth,
        height=0.65\textheight,
        keepaspectratio
    ]{figures/figure_folder/Collage_WACV.pdf}
    \vspace{-0.2cm}
    \captionof{figure}{\small Qualitative results generated by \ours{}.}
    \label{fig:SuppFigure}
\end{strip}
}

\newcommand{\introfig}{
\begin{figure}[]
    \centering
    \includegraphics[
        width=0.99\columnwidth
    ]{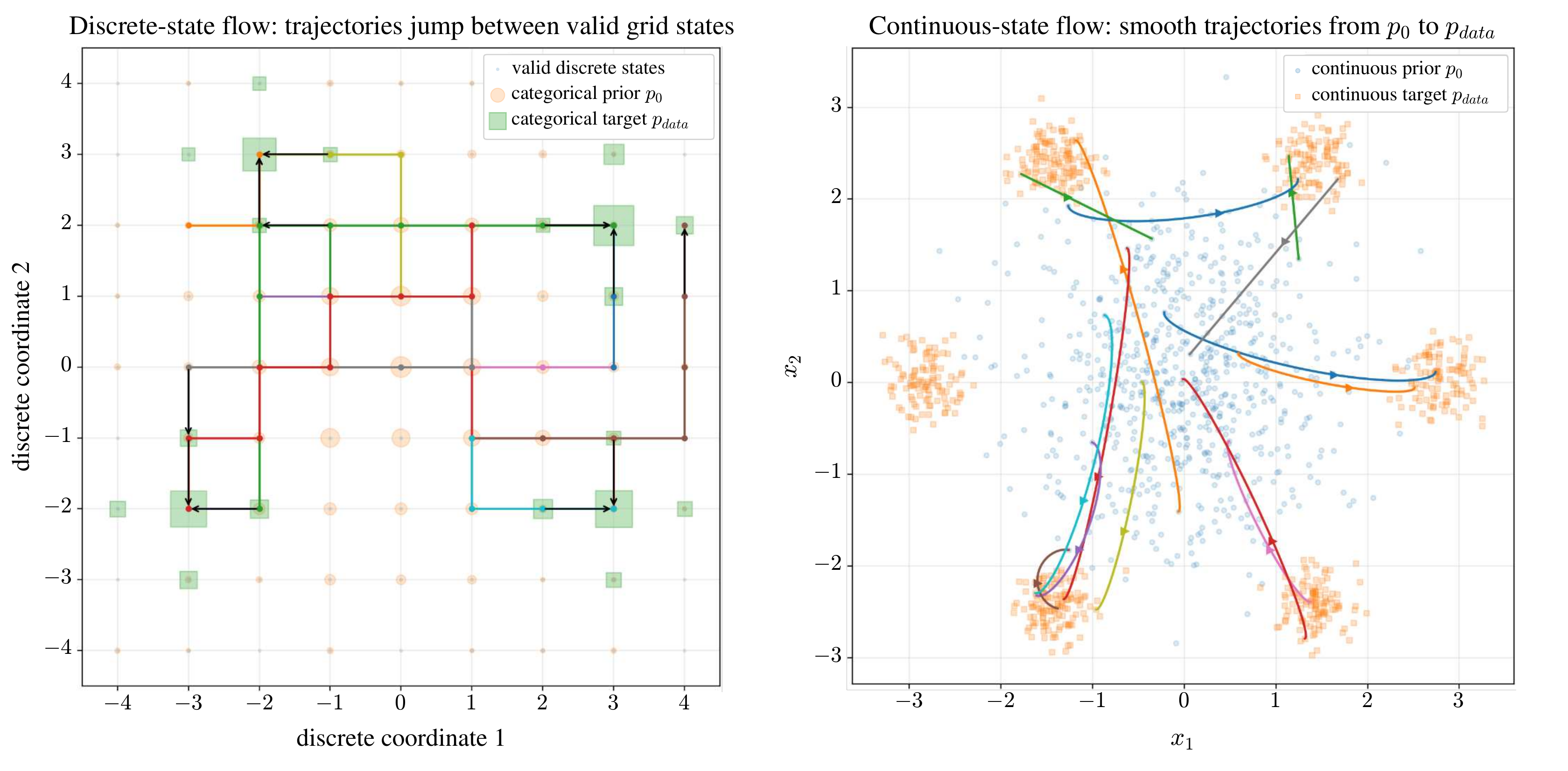}
    \vspace{-0.4cm}
    \caption{ \small
        Discrete vs. continuous generative transport: discrete models jump between categorical states, while Flow Matching follows smooth ODE trajectories from prior to target.
    }
    \label{fig:discrete_continuous_flow}
    \vspace{-0.4cm}
\end{figure}
}

\newcommand{\TeaserFigure}{
\begin{figure*}[t]
    \centering
    \includegraphics[width=\linewidth]{figures/figure_folder/teasser_dig_2.pdf}

    \caption{\textbf{Teaser Figure}}
    \label{fig:TeaserFigure}

\end{figure*}
}

\newcommand{\TableAblation}{
\begin{table*}[t!]
\centering
\scriptsize
\caption{\small Ablation study on the impact of distillation, teacher models, and parameter-efficient adaptation. Subscript values denote changes relative to the Main Baseline.}
\label{tab:ablation_study}
\vspace{-0.3cm}
\resizebox{0.99\textwidth}{!}{
\begin{tabular}{lcccccc}
\toprule

\multirow{3}{*}{\textbf{Variants}}
& \multicolumn{4}{c}{\textbf{Image Generation}}
& \multicolumn{2}{c}{\textbf{Semantic Understanding}} \\

\cmidrule(lr){2-5}
\cmidrule(lr){6-7}

& \textbf{GenEval}
& \textbf{COCO-30K}
& \multicolumn{2}{c}{\textbf{Aesthetic-Eval}}
& \textbf{MMBench}
& \textbf{MMMU} \\

\cmidrule(lr){2-2}
\cmidrule(lr){3-3}
\cmidrule(lr){4-5}
\cmidrule(lr){6-6}
\cmidrule(lr){7-7}

& Accuracy $\uparrow$
& FID $\downarrow$
& CLIP Score $\uparrow$
& Aesthetic Score $\uparrow$
& Accuracy $\uparrow$
& Accuracy $\uparrow$ \\

\midrule

\rowcolor{gray!10}
\multicolumn{7}{c}{
\textit{Main Baseline (Qwen as teacher, LoRA adapters, and UniDFlow as base unified model)}
} \\

Main Baseline
& 0.92
& 8.92
& 38.23
& 6.21
& 90.92
& 74.26 \\

\rowcolor{gray!10}
\multicolumn{7}{c}{
\textit{Impact of Distillation}
} \\

No Teacher Model
& 0.89$_{-0.03}$
& 10.69$_{+1.77}$
& 36.31$_{-1.92}$
& 5.58$_{-0.63}$
& 89.14$_{-1.78}$
& 73.28$_{-0.98}$ \\

\rowcolor{gray!10}
\multicolumn{7}{c}{
\textit{Impact of Different Teacher Models}
} \\

Flux 1.0~\cite{flux1}
& 0.91$_{\textcolor{blue}{-0.01}}$
& 9.91$_{\textcolor{blue}{+0.99}}$
& 38.05$_{\textcolor{blue}{-0.18}}$
& 6.08$_{\textcolor{blue}{-0.13}}$
& 90.23$_{\textcolor{blue}{-0.69}}$
& 74.24$_{\textcolor{blue}{-0.02}}$ \\

Flux 2 4B~\cite{flux-2-2025}
& 0.92$_{\textcolor{blue}{+0.00}}$
& 10.45$_{\textcolor{blue}{+1.53}}$
& 38.13$_{\textcolor{blue}{-0.10}}$
& 6.08$_{\textcolor{blue}{-0.13}}$
& 90.55$_{\textcolor{blue}{-0.37}}$
& 74.39$_{\textcolor{blue}{+0.13}}$ \\

Flux 2 9B~\cite{flux-2-2025}
& 0.92$_{\textcolor{blue}{+0.00}}$
& 8.92$_{\textcolor{blue}{+0.00}}$
& 38.17$_{\textcolor{blue}{-0.06}}$
& 6.17$_{\textcolor{blue}{-0.04}}$
& 90.58$_{\textcolor{blue}{-0.34}}$
& 74.67$_{\textcolor{blue}{+0.41}}$ \\

Flux 2 32B~\cite{flux-2-2025}
& 0.91$_{\textcolor{blue}{-0.01}}$
& 8.33$_{\textcolor{blue}{-0.59}}$
& 38.34$_{\textcolor{blue}{+0.11}}$
& 6.24$_{\textcolor{blue}{+0.03}}$
& 90.41$_{\textcolor{blue}{-0.51}}$
& 74.17$_{\textcolor{blue}{-0.09}}$ \\

\rowcolor{gray!10}
\multicolumn{7}{c}{
\textit{Full Fine-Tuning vs.\ LoRA}
} \\

No Teacher Model and No Adapters
& 0.88$_{\textcolor{blue}{-0.04}}$
& 9.92$_{\textcolor{blue}{+1.00}}$
& 37.11$_{\textcolor{blue}{-1.12}}$
& 5.87$_{\textcolor{blue}{-0.34}}$
& 86.35$_{\textcolor{blue}{-4.57}}$
& 70.01$_{\textcolor{blue}{-4.25}}$ \\

No Adapters
& 0.90$_{\textcolor{blue}{-0.02}}$
& 8.76$_{\textcolor{blue}{-0.16}}$
& 38.52$_{\textcolor{blue}{+0.29}}$
& 6.00$_{\textcolor{blue}{-0.21}}$
& 86.01$_{\textcolor{blue}{-4.91}}$
& 69.28$_{\textcolor{blue}{-4.98}}$ \\

\bottomrule
\end{tabular}
}
\end{table*}
}

\newcommand{\TableGeneration}{
\begin{table*}[t]
\centering
\scriptsize
\caption{\small Image generation performance, model size, and inference time.
Subscript values denote the score change of each FlowShift model relative
to its corresponding DFM source model.}
\label{tab:image_generation}
\vspace{-0.3cm}
\resizebox{0.99\textwidth}{!}{
\begin{tabular}{lcccccccc}
\toprule

\multirow{2}{*}{\textbf{Baselines}}
& \multirow{2}{*}{\textbf{Params (B)}}
& \multirow{2}{*}{\textbf{Inf. Time (s)}}
& \textbf{GenEval}~\cite{ghosh2023geneval}
& \multicolumn{2}{c}{\textbf{COCO-30K}~\cite{lin2014microsoft}}
& \multicolumn{3}{c}{\textbf{Aesthetic-Eval}~\cite{zhang2025diffusion}} \\

\cmidrule(lr){4-4}
\cmidrule(lr){5-6}
\cmidrule(lr){7-9}

& & 
& Accuracy $\uparrow$
& FID $\downarrow$
& DINOv2-FID $\downarrow$
& FID $\downarrow$
& CLIPScore $\uparrow$
& Aesthetic Score $\uparrow$ \\

\midrule

\rowcolor{gray!10}
\multicolumn{9}{c}{
\textit{Continuous Flow Matching (CFM) based Native Image Generation Models}
} \\

Flux1~\cite{flux1}
& 14
& 16
& 0.66
& 9.15
& 10.43
& 39.41
& 33.42
& 5.51 \\

FLUX.2 [klein] 9B~\cite{flux-2-2025}
& 9
& 12
& 0.75
& 8.92
& 9.17
& 38.47
& 35.41
& 6.28 \\

Qwen-Image RL~\cite{wu2025qwen}
& 22
& 21
& 0.88
& 8.55
& 9.32
& 39.09
& 33.98
& 6.43 \\

Flux-2 32B~\cite{flux-2-2025}
& 32
& 33
& 0.89
& 7.12
& 7.77
& 33.14
& 39.21
& 7.56 \\

Z-Image~\cite{cai2025z}
& 6
& 8
& 0.87
& 8.18
& 8.28
& 37.79
& 36.29
& 5.85 \\

\rowcolor{gray!10}
\multicolumn{9}{c}{
\textit{Continuous Flow Matching (CFM) based Unified Models}
} \\

OmniGen2~\cite{wu2026omnigen2}
& $3{+}4$
& 7
& 0.85
& 8.89
& 9.48
& 40.08
& 37.32
& 5.18 \\

BAGEL~\cite{deng2025emerging}
& $7{+}7$
& 12
& 0.88
& 9.32
& 9.19
& 41.14
& 36.61
& 5.92 \\

MammothModa-2~\cite{shen2025mammothmoda2}
& $8{+}3{+}2$
& 11
& 0.87
& 10.02
& 9.95
& 40.44
& 35.92
& 6.05 \\

\rowcolor{gray!10}
\multicolumn{9}{c}{
\textit{Discrete Flow Matching (DFM) based Unified Models}
} \\

MUDDIT~\cite{shi2025muddit}
& 2
& 5
& 0.67
& 10.11
& 16.68
& 50.16
& 33.36
& 3.15 \\

FUDOKI~\cite{wang2026fudoki}
& 2
& 7
& 0.81
& 16.16
& 13.29
& 48.93
& 38.77
& 4.42 \\

UniDFlow-4B~\cite{susladkar2026best}
& 4
& 5
& 0.95
& 13.33
& 15.23
& 44.78
& 38.01
& 4.43 \\

\rowcolor{gray!10}
\multicolumn{9}{c}{
\textit{DFM $\rightarrow$ CFM Unified Models}
} \\

\ours{} (FUDOKI)
& 2
& 8
& $0.77_{\scriptscriptstyle \textcolor{blue}{-0.04}}$
& $9.29_{\scriptscriptstyle \textcolor{blue}{-6.87}}$
& $10.15_{\scriptscriptstyle \textcolor{blue}{-3.14}}$
& $38.42_{\scriptscriptstyle \textcolor{blue}{-10.51}}$
& $37.17_{\scriptscriptstyle \textcolor{blue}{-1.60}}$
& $5.92_{\scriptscriptstyle \textcolor{blue}{+1.50}}$ \\

\ours{} (UniDFlow-4B)
& 4
& 7
& $0.92_{\scriptscriptstyle \textcolor{blue}{-0.03}}$
& $8.92_{\scriptscriptstyle \textcolor{blue}{-4.41}}$
& $9.53_{\scriptscriptstyle \textcolor{blue}{-5.70}}$
& $38.32_{\scriptscriptstyle \textcolor{blue}{-6.46}}$
& $38.23_{\scriptscriptstyle \textcolor{blue}{+0.22}}$
& $6.21_{\scriptscriptstyle \textcolor{blue}{+1.78}}$ \\

\bottomrule
\end{tabular}
}
\end{table*}
}

\newcommand{\TableUnderstanding}{
\begin{table*}[t!]
\centering
\scriptsize
\setlength{\tabcolsep}{8pt} 
\caption{\small Multimodal understanding performance. Subscript values beside
FlowShift models denote the average percentage of source-model
performance retained after conversion.}
\label{tab:understanding_results}
\vspace{-0.3cm}
\resizebox{0.99\linewidth}{!}{%
\begin{tabular}{lccccc}
\toprule

\textbf{Baselines}
& \textbf{MMBench~\cite{liu2024mmbench}}
& \textbf{MMMU~\cite{yue2024mmmu}}
& \textbf{MM-VET~\cite{yu2023mm}}
& \textbf{MathVista~\cite{lu2024mathvista}}
& \textbf{MMVP~\cite{tong2024eyes}} \\

\midrule

\rowcolor{gray!10}
\multicolumn{6}{c}{
\textit{Continuous Flow Matching (CFM) based Unified Models}
} \\

BAGEL~\cite{deng2025emerging}
& 85.00
& 55.30
& 67.20
& 73.10
& 69.30 \\

OmniGen2~\cite{wu2026omnigen2}
& 79.10
& 53.10
& 61.80
& 69.42
& 68.26 \\

MammothModa-2~\cite{shen2025mammothmoda2}
& 86.60
& 71.23
& 79.40
& 81.80
& 77.50 \\

\rowcolor{gray!10}
\multicolumn{6}{c}{
\textit{Discrete Flow Matching (DFM) based Unified Models}
} \\

MUDDIT~\cite{shi2025muddit}
& 55.48
& 49.87
& 40.14
& 70.15
& 66.66 \\

\rowcolor{gray!10}
\multicolumn{6}{c}{
\textit{Semantic Preservation under DFM $\rightarrow$ CFM Conversion}
} \\

FUDOKI (Source)~\cite{wang2026fudoki}
& 81.27
& 58.01
& 65.23
& 74.17
& 67.72 \\

\ours{} (FUDOKI)$_{\scriptscriptstyle \textcolor{blue}{99.73\%}}$
& 81.19
& 58.00
& 64.91
& 73.29
& 68.01 \\

\cmidrule(lr){1-6}

UniDFlow-4B (Source)~\cite{susladkar2026best}
& 91.20
& 74.50
& 82.70
& 85.90
& 80.20 \\

\ours{} (UniDFlow-4B)$_{\scriptscriptstyle \textcolor{blue}{99.54\%}}$
& 90.92
& 74.26
& 82.40
& 85.20
& 79.80 \\

\bottomrule
\end{tabular}%
}
\vspace{-0.3cm}
\end{table*}
}

\begin{abstract}
\vspace{-17pt}

While discrete tokenization in Unified Multimodal Models (UMMs) establishes strong semantic reasoning, it introduces a strict quantization bottleneck that degrades fine-grained visual details, often resulting in structurally accurate but artifact-heavy images. We introduce \oursbold{}, a parameter-efficient post-training framework that bypasses this bottleneck by converting pre-trained discrete UMMs into continuous rectified-flow generators. Instead of discarding the backbone's semantic priors, \ours{} replaces discrete visual interfaces with continuous latent projections and applies LoRA to the attention layers. To efficiently boost visual quality without semantic drift, we propose Dual-Objective Velocity Vector Distillation (DOVD), which aligns the model with a high-fidelity continuous teacher and ground-truth continuous trajectories. This approach drastically improves perceptual quality while preserving prompt adherence and reasoning. When applied to leading discrete UMMs, \ours{} significantly improves COCO-30K FID and aesthetic scores, while retaining over $99.5\%$ of multimodal understanding capabilities. Ablations confirm DOVD drives generation quality. 
Ultimately, \ours{} achieves high-fidelity continuous generation without sacrificing pre-trained semantic grounding.

\end{abstract}
    
\section{Introduction}
\label{sec:intro}

\introfig

The landscape of multimodal AI is shifting toward Unified Multimodal Models (UMMs) that jointly handle understanding and generation within a single framework~\cite{deng2025emerging, chen2025janus, team2024chameleon, xie2025show, wang2024emu3, shen2026latteflow, yu2026elsa3d, yu2025core3d}. Practical UMMs require strong semantic grounding and high-quality visual synthesis~\cite{rewardflow2026,venkatesh2026ravel}, leading to two main architectural paradigms. Models such as Chameleon~\cite{team2024chameleon} and Show-o~\cite{xie2025show} map visual data into discrete tokens, integrating naturally with the autoregressive next-token prediction paradigm of large language models. Conversely, hybrid models like Transfusion~\cite{zhou2025transfusion} combine autoregressive text modeling with continuous image generation.\looseness-1

While discrete image representations integrate naturally with language models and preserve robust semantic reasoning, their reliance on quantization and categorical prediction fundamentally restricts fine-grained reconstruction~\cite{zheng2025rethinking, li2025discrete}. At a fixed codebook size and token resolution, multiple continuous visual features may map to the same discrete representation, limiting the exact preservation of fine-grained appearance variations such as texture and local shading. Consequently, models fail to learn the smooth gradients required for photorealism, producing structurally accurate but artifact-heavy images. Although expanding codebook sizes mitigates quantization error~\cite{li2025discrete}, it triggers vocabulary explosion and exacerbates the data-hungry nature of visual modeling, offering diminishing returns compared to continuous approaches~\cite{shukor2025scaling, zhou2025transfusion}.

Conversely, continuous latent flows offer smoother visual geometry but commonly require massive retraining from scratch to establish semantic understanding~\cite{yang2025survey, zhao2025unified}. Their unbounded, high-dimensional spaces make learning strict semantic constraints (e.g., exact counting, rigid spatial relations) mathematically difficult, increasing the risk of out-of-distribution hallucinations. Furthermore, continuous models heavily optimized for visual appeal, often via Direct Preference Optimization (DPO)~\cite{rafailov2023direct} frequently suffer from reward hacking, thus sacrificing faithful instruction adherence for scalar aesthetic rewards~\cite{labs2025flux}.

Figure~\ref{fig:discrete_continuous_flow} contrasts discrete and continuous state generative transport. In the discrete setting, samples lie on a finite or countable support and evolve through categorical transitions, yielding jump-like trajectories over quantized spaces. In contrast, continuous Flow Matching transports samples smoothly via a time-dependent velocity field $v_{\theta}(x,t)$ in a Euclidean latent space, avoiding the quantization bottleneck and supporting fine-grained visual variation~\cite{liu2022flow}, finding applications in 3D generation, video generation, and video editing~\cite{shen2026phantom,yu2026silsa,juvekar2026editvid,xiang2025structured,wan2025}.
Motivated by this distinction, we propose a post-training framework \ours{} that converts a pre-trained discrete backbone from categorical token prediction to continuous latent-vector regression while preserving its semantic priors. We further introduce Dual-Objective Velocity Vector Distillation (DOVD), which transfers transport dynamics from a high-fidelity continuous teacher while maintaining the student’s semantic controllability. This enables improved high-frequency rendering and aesthetic fidelity without requiring full retraining or substantial parameter scaling. Our contributions are:\looseness-1

\begin{itemize}
\item We propose \oursbold{}, a parameter-efficient post-training framework that converts a pre-trained discrete Unified Multimodal Model (UMM) into a continuous rectified-flow generator without retraining the backbone, thereby preserving the semantic and compositional priors learned during discrete multimodal pretraining.

\item We introduce \textbf{Dual-Objective Velocity Vector Distillation (DOVD)}, a distillation objective that operates over a shared continuous VAE latent space. By jointly optimizing for ground-truth flow matching and continuous velocity guidance from a frozen teacher model, DOVD overcomes the quantization bottleneck to recover fine-grained structural and photometric details.

\item We validate \ours{} across multiple discrete UMM backbones and continuous teachers, demonstrating consistent improvements in perceptual quality with minimal semantic degradation. Across the evaluated backbones, COCO-30K FID improves by $33.1\%$--$42.5\%$ and the aesthetic score by $1.50$--$1.78$ points, while more than $99.5\%$ of the original multimodal-understanding performance is retained across five benchmarks.
\end{itemize}
\section{Related Work}
\label{sec:related_work}

\MethodologyFigure

\noindent\textbf{{Unified Multimodal Models}} 
(UMMs) have evolved through distinct architectural paradigms to jointly support text and image generation within a single framework: 

\noindent \textbf{1. Discrete autoregressive models.} By mapping modalities into a shared discrete vocabulary, early models enabled unified training via standard next-token prediction. Chameleon~\cite{team2024chameleon} and Emu3~\cite{wang2024emu3} pioneered this early-fusion of interleaved sequences, which Janus later refined by decoupling the visual encoding pathways for understanding and generation. Furthermore, Unified-IO 2~\cite{lu2024unified} expanded this discrete paradigm to encompass audio, actions, and structured outputs. 

\noindent \textbf{2. Discrete-diffusion models.} To address the slow, unidirectional nature of autoregressive decoding, subsequent models adopted discrete diffusion. Show-o~\cite{xie2025show} combines autoregressive text and discrete diffusion over visual tokens within a single transformer. UniDisc~\cite{swerdlow2025unified}  extended this by employing discrete diffusion as a joint formulation for both text and images, allowing for bidirectional denoising. MMaDA~\cite{yang2026mmada}  similarly applies a modality-agnostic discrete diffusion objective to unify textual reasoning and visual synthesis. More recently, FUDOKI~\cite{wang2026fudoki} replaces autoregressive modeling with kinetic-optimal discrete flow matching, enabling unified understanding and iterative image generation. UniDFlow~\cite{susladkar2026best} further unifies understanding, generation, and editing using task-specific adapters and preference alignment to preserve reasoning and controllability.~\looseness-1

\noindent \textbf{3. Continuous AR–diffusion.} To achieve higher visual fidelity, the field is rapidly transitioning from shared discrete spaces to continuous visual dynamics. Transfusion~\cite{zhou2025transfusion} pairs text autoregression with continuous image diffusion for superior scaling, while MonoFormer~\cite{zhao2024monoformer} leverages a shared transformer for causal text and bidirectional continuous image diffusion. Similarly, recent architectures are directly upgrading their discrete predecessors: JanusFlow~\cite{ma2025janusflow} replaces the discrete autoregression of Janus~\cite{wu2025janus} with continuous rectified flow, and Show-o2~\cite{xie2026show} evolves Show-o from discrete diffusion to continuous flow matching over VAE representations.

While the progression to continuous hybrids highlights high-fidelity generation, training them from scratch demands exorbitant compute and vast datasets to establish semantic alignment. As an efficient alternative, our method \ours{} post-trains existing discrete models to capture aesthetic benefits of continuous flow-matching, bypassing the massive retraining costs of native hybrids while fully preserving the backbone's original semantic properties.

\subsection{Quantization Bottleneck} 

The shift toward continuous hybrids stems from the severe information loss inherent in discrete tokenization pipelines established by VQ-VAE~\cite{van2017neural} and VQGAN~\cite{esser2021taming}, which map continuous images into finite codebooks. To mitigate the resulting quantization error, recent advancements have focused on enhancing tokenizer capacity~\cite{susladkar2026pyratok}. MAGVIT-v2~\cite{yu2024language}  introduced lookup-free quantization and massively expanded vocabularies to make discrete generation competitive with continuous diffusion. Further efforts like Finite Scalar Quantization~\cite{mentzer2024finite}  address codebook collapse.

Despite tokenizer improvements, discretization fundamentally throttles generative quality due to a granularity conflict: TokenFlow~\cite{qu2025tokenflow} shows that while semantics benefit from discrete compression, generation requires continuous information. Consequently, VTBench~\cite{lin2025vtbench} confirms discrete tokenizers lag behind continuous VAEs in reconstructing fine textures and structural integrity. Furthermore, Transfusion~\cite{zhou2025transfusion} proves continuous diffusion scales significantly better than quantized next-token prediction. 
By employing DOVD to transfer smooth trajectory dynamics, our framework bypasses this bottleneck, recovering the fine-grained details and high-frequency textures that discrete vocabularies erase.

\section{Method}
\label{sec:method}
 
\subsection{Overview}
\label{subsec:overview}

Discrete models offer strong semantics but lack perceptual fidelity due to quantization and categorical training objectives. Conversely, continuous models achieve high fidelity but demand massive training for semantic alignment and often sacrifice instruction adherence when optimized for aesthetics. While existing continuous hybrids bridge this gap, resolving the trade-off by training from scratch remains computationally prohibitive. 

We address this with \oursbold{}, a post-training framework that converts a pretrained discrete unified model into a continuous rectified-flow generator without updating pretrained weights (Fig.~\ref{fig:MethodologyFigure}). \ours{} comprises three components: (1) replacing discrete visual interfaces with continuous ones to enable velocity regression in a VAE latent space shared with a frozen teacher; (2) applying LoRA to attention projections while keeping all backbone and teacher parameters frozen; and (3) Dual-Objective Velocity Vector Distillation (DOVD), which combines flow-matching loss with teacher velocity distillation.

\subsection{Problem Formulation}
\label{subsec:formulation}

Let $\mathcal{M}_{\theta}$ be a pretrained UMM with frozen parameters $\theta$, which encodes an image $\mathbf{x}$ into $N$ categorical tokens $\mathbf{s}\!=\!\mathcal{Q}(\mathbf{x})\in\{1,\dots,K\}^{N}$ using a tokenizer $\mathcal{Q}$ of codebook size $K$.
For a model that generates images by autoregressively predicting and decoding these tokens, the objective can be written as
\begin{equation}
\setlength{\abovedisplayskip}{6pt}
\setlength{\belowdisplayskip}{6pt}
\setlength{\abovedisplayshortskip}{6pt}
\setlength{\belowdisplayshortskip}{6pt}
    \mathcal{L}_{\mathrm{AR}}(\theta) \!=\! -\,\mathbb{E}_{(\mathbf{x},\mathbf{y})}\Big[\textstyle\sum_{i\!=\!1}^{N}\log p_{\theta}\big(s_{i}\mid\mathbf{s}_{<i},\mathbf{y}\big)\Big],
    \label{eq:ar}
\end{equation}
where $\mathbf{y}$ denotes the prompt.
This categorical formulation has two limitations: high-frequency content mapped to codebook centroids is unrecoverable by any decoder, and cross-entropy loss penalizes all incorrect codes equally.

We instead adopt the rectified-flow formulation~\cite{liu2022flow}. Given a VAE with encoder $\mathcal{E}_{\mathrm{VAE}}$, decoder $\mathcal{D}_{\mathrm{VAE}}$, and latents $\mathbf{z}_{0}\!=\!\mathcal{E}_{\mathrm{VAE}}(\mathbf{x})\in\mathbb{R}^{c\times h\times w}$, noise and data are connected by the linear interpolant $\mathbf{z}_{t}\!=\!(1-t)\,\mathbf{z}_{0}+t\,\boldsymbol{\epsilon}$.

where $\boldsymbol{\epsilon}\!\sim\!\mathcal{N}(\mathbf{0},\mathbf{I})$ and $t\in[0,1]$ (with $t\!=\!0$ as data and $t\!=\!1$ as noise). The conditional velocity is directly given by $\dot{\mathbf{z}}_{t}\!=\!\boldsymbol{\epsilon}-\mathbf{z}_{0}$. By regressing this target, the model learns the marginal velocity field, denoted as $\mathbf{v}^{\star}_{t}(\mathbf{z}\mid\mathbf{y})\!=\!\mathbb{E}[\boldsymbol{\epsilon}-\mathbf{z}_{0}\mid\mathbf{z}_{t}\!=\!\mathbf{z},\mathbf{y}]$. This field defines an Ordinary Differential Equation (ODE), $\mathrm{d}\mathbf{z}/\mathrm{d}t\!=\!\mathbf{v}^{\star}_{t}(\mathbf{z}\mid\mathbf{y})$. To generate samples, this ODE is integrated from $t\!=\!1$ to $t\!=\!0$ and the final latent is decoded via $\hat{\mathbf{x}}\!=\!\mathcal{D}_{\mathrm{VAE}}(\mathbf{z}_{0})$. Unlike Eq.~\eqref{eq:ar}, this Euclidean regression provides gradients directly proportional to the prediction error.

We formulate our problem as follows: given a frozen discrete unified backbone $\mathcal{M}_\theta$, a frozen rectified-flow teacher $\mathcal{T}_{\xi}$, and an image-text dataset $\mathcal{D}$, we learn a lightweight parameter set $\phi$ (where $\vert{}\phi\vert{}\!\ll\!\vert{}\theta\vert{}$) such that the augmented model $\mathcal{M}_{\phi}(\mathbf{z}_{t},t,\mathbf{y})$ (i) defines a continuous velocity field with improved perceptual fidelity; (ii) inherits semantic and compositional control via the backbone's native language pathway; and (iii) requires only parameter-efficient post-training, bypassing full multimodal pretraining.

Hereafter, $\mathcal{M}_{\phi}$ denotes the backbone $\mathcal{M}_{\theta}$ equipped with the trainable modules in $\phi$, and the frozen parameters $\theta$ are left implicit. The next subsection presents how $\mathcal{M}_\phi$ is constructed and trained.
 
\subsection{\textbf{\ours{}}}
\TableGeneration

\TableUnderstanding

\label{subsec:our_approach}

\ours{} utilizes two strictly frozen pretrained networks: the unified backbone $\mathcal{M}_{\theta}$, which forms the frozen core of the student $\mathcal{M}_{\phi}$ and the rectified-flow teacher $\mathcal{T}_{\xi}$. The trainable parameters ($\phi$) are a continuous input projector $\mathbf{W}_{\mathrm{in}}$, a timestep layer $\mathrm{MLP}_{t}$, attention LoRA adapters, and a velocity output head $\mathbf{W}_{\mathrm{out}}$.

For an image-prompt pair $(\mathbf{x}, \mathbf{y}) \in \mathcal{D}$, the teacher's frozen VAE encoder acts as a shared interface, extracting the continuous clean latent $\mathbf{z}_{0}\!\!=\!\!\mathcal{E}_{\mathrm{VAE}}(\mathbf{x})$. Following, noise is added to produce $\mathbf{z}_{t}$, which is then passed to both the student ($\mathcal{M}_{\phi}$) and teacher ($\mathcal{T}_{\xi}$). Sharing the exact clean latent, noise, and timestep ensures both networks predict velocities at the identical trajectory point.

Simultaneously, each model processes the prompt $\mathbf{y}$ through its native text encoder  to yield conditioning signals $\mathbf{Y}_{S}$ (student) \textbf{and} $\mathbf{Y}_{T}$ (teacher). We share VAE latents to enable pointwise comparison of the teacher’s and student’s velocity predictions. Text encoders remain separate because each model requires its native conditioning distribution; cross-model embeddings could corrupt supervision and weaken the student’s semantic grounding.

Of the two networks, only the student requires architectural modification; the teacher is inherently a continuous velocity network that directly predicts $\mathbf{v}_{T}\!=\!\mathcal{T}_{\xi}(\mathbf{z}_{t},t,\mathbf{y})$ from the shared noisy latent. Since the frozen student backbone natively expects and predicts discrete tokens, we adapt its forward pass at three distinct points: the input interface, attention computation, and output interface.

\noindent \textbf{Input Adaptation.} At input, $\mathbf{z}_{t}$ is patchified (size $p$) and projected to model width $d$ via trainable layer $\mathbf{W}_{\mathrm{in}}$, producing $N\!=\!hw/p^{2}$ continuous tokens that match the discrete sequence length of Eq.~\eqref{eq:ar}. By mapping these to the original discrete positional indices, we directly reuse the backbone's pretrained spatial priors. Simultaneously, timestep $t$ is injected by adding its sinusoidal embedding, transformed by the two-layer network $\mathrm{MLP}_{t}$, to every visual token.\looseness-1

\noindent \textbf{Backbone Adaptation.} The visual tokens then enter the frozen backbone together with $\mathbf{Y}_{S}$. Inside $\mathcal{M}_{\theta}$, we attach LoRA adapters~\cite{hu2021lora} to the query, key, value, and output projections of every self-attention layer. For layer $\ell$ and projection type $a\in\{Q,K,V,O\}$, the adapted projection is 
$\widetilde{\mathbf{W}}^{\ell}_{a} \!=\!
\mathbf{W}^{\ell}_{a} + \frac{\alpha}{r} \mathbf{B}^{\ell}_{a} \mathbf{A}^{\ell}_{a},$ where $\mathbf{W}^{\ell}_{a}$ is frozen, $\mathbf{A}^{\ell}_{a}$ and $\mathbf{B}^{\ell}_{a}$ are trainable low-rank matrices, $r$ is the LoRA rank, and $\alpha$ is the LoRA scaling coefficient. 
We restrict adaptation to attention projections because the representation transition primarily changes how continuous visual tokens interact with one another and with the textual condition. Updates to the query and key projections adapt the attention geometry, while updates to the value and output projections adapt how continuous visual information is aggregated and propagated. Freezing the feed-forward layers and remaining backbone parameters preserves pretrained semantic representations and reduces semantic drift.

\noindent \textbf{Output Adaptation.} At the output, the hidden states at the visual positions are mapped by the linear head $\mathbf{W}_{\mathrm{out}}$ and unpatchified into the student prediction $\mathbf{v}_{S}\!=\!\mathcal{M}_{\phi}(\mathbf{z}_{t},t,\mathbf{y})$, replacing the original softmax head over the codebook. This completes the architecture. Trainable parameters are therefore given by\looseness-1
\begin{equation}
\setlength{\abovedisplayskip}{6pt}
\setlength{\belowdisplayskip}{6pt}
\setlength{\abovedisplayshortskip}{6pt}
\setlength{\belowdisplayshortskip}{6pt}
    \phi
    \!=\!
    \left\{
        \mathbf{W}_{\mathrm{in}},
        \mathbf{W}_{\mathrm{out}},
        \mathrm{MLP}_{t},
        \left\{
            \mathbf{A}^{\ell}_{a},
            \mathbf{B}^{\ell}_{a}
        \right\}_{\ell,a}
    \right\}.
    \label{eq:trainable_parameters}
\end{equation}
All parameters of the source text-conditioning pathway, unified backbone, teacher text encoder, teacher transformer, and VAE remain frozen.

\noindent \textbf{Training Objective:} With both networks predicting velocity vectors over the same noisy latent, relying on a single training signal is suboptimal. Standard conditional flow matching ties the student to the target data distribution, but recovering high-frequency generation behavior purely from data is highly inefficient. Conversely, pure teacher distillation efficiently transfers continuous rendering dynamics but risks transferring the teacher's instruction-adherence errors.

Dual-Objective Velocity Vector Distillation (DOVD) therefore jointly optimizes the student against the analytic rectified-flow target and the frozen teacher prediction. With expectations taken over
$(\mathbf{x},\mathbf{y})\!\sim\!\mathcal{D}$, $\boldsymbol{\epsilon}\!\sim\!\mathcal{N}(\mathbf{0},\mathbf{I})$, and $t\!\sim\!\pi_{t}$, the objective is
\begin{align}
    \mathcal{L}_{\mathrm{FM}}(\phi) &\!=\! \mathbb{E}\Big[\big\lVert \mathbf{v}_{S} - (\boldsymbol{\epsilon}-\mathbf{z}_{0})\big\rVert_{2}^{2}\Big],
    \label{eq:fm}\\
    \mathcal{L}_{\mathrm{VD}}(\phi) &\!=\! \mathbb{E}\Big[\big\lVert \mathbf{v}_{S} - \sg[\mathbf{v}_{T}]\big\rVert_{2}^{2}\Big],
    \label{eq:vd}\\
    \mathcal{L}_{\mathrm{DOVD}}(\phi) &\!=\! \mathcal{L}_{\mathrm{FM}}(\phi) + \lambda\,\mathcal{L}_{\mathrm{VD}}(\phi),
    \label{eq:total}
\end{align}
where $\sg[\cdot]$ denotes stop-gradient, $\lambda$ weights the distillation term.
The two objectives serve different purposes. The Flow Matching (FM) loss, $\mathcal{L}_{\mathrm{FM}}$, preserves direct supervision from the prompt--image training pair and prevents the student from merely imitating the teacher. In contrast, the Velocity Distillation (VD) loss,  $\mathcal{L}_{\mathrm{VD}}$, regularizes the student toward the smooth velocity field learned by a high-capacity continuous generator. Together, they help the student inherit continuous rendering without relying solely on the teacher for visual or semantic supervision.

\section{Experiments}
\label{sec:experimental}
\UniDFlowFigure

We train \ours{} on a curated corpus of approximately 6 million image-text pairs sourced from LAION-20M Recap, OpenGPT-4o-Image, and DRAGON. To ensure fair and rigorous comparisons against all discrete and continuous baselines, we maintain strictly consistent evaluation protocols across all models, utilizing identical evaluation prompt sets, standardized sampling steps, and matching generation resolutions. During inference, generations employ an Euler discrete scheduler with a uniform classifier-free guidance scale of 8.0. Comprehensive details regarding the training data, evaluation benchmarks, metrics, and implementation settings are provided in Appendix ~\ref{sec:supp_experimental_sec}.

\section{Quantitative Results}
\label{sec:quantitative}

\subsection{Image Generation}

Table~\ref{tab:image_generation} compares \ours{}-adapted models (listed under \emph{DFM $\rightarrow$ CFM Unified Models}) against three baselines: native continuous generators (\emph{CFM based Native Image Generation Models}), scratch-trained continuous UMMs (\emph{CFM based Unified Models}), and discrete UMMs (\emph{DFM based Unified Models}). Results show \ours{} effectively balances semantic control (prompt adherence) and perceptual fidelity (visual quality).

As baselines demonstrate, discrete unified models excel in compositional fidelity (e.g., UniDFlow-4B~\cite{susladkar2026best} scores $0.95$ on GenEval) but suffer on perceptual metrics due to quantization-induced high-frequency loss. Conversely, native continuous models leverage smooth latent geometry for superior visuals. For instance, FLUX-2 32B~\cite{flux-2-2025} achieves the best COCO-30K FID ($7.12$) and Aesthetic scores ($7.56$) yet struggles with complex instructions (GenEval $0.89$).

When observing recent continuous unified models trained from scratch, such as OmniGen2~\cite{wu2026omnigen2} and BAGEL~\cite{deng2025emerging}, the empirical results highlight the difficulty of optimizing textual reasoning and continuous generation simultaneously. 
Without a discrete semantic foundation, the data shows these models compromise on both fronts: they fail to match the rigorous logical grounding of discrete models (with GenEval scores hovering between $0.85$ and $0.88$) and fall short of the pristine visual polish of native continuous generators (yielding Aesthetic scores below $6.0$).

Rather than relearning these capabilities from scratch, \ours{} transfers a learned discrete generation policy into a continuous space. By restricting adaptation to visual generation and leveraging teacher trajectory supervision, the model preserves its semantic backbone while improving perceptual modeling. Bypassing the quantization bottleneck yields substantial improvements in visual fidelity across architectures: FUDOKI~\cite{wang2026fudoki} improves COCO-30K FID from $16.16$ to $9.29$ and Aesthetic Score from $4.42$ to $5.92$, while UniDFlow-4B improves FID from $13.33$ to $8.92$ and Aesthetic Score from $4.43$ to $6.21$. Compositional performance remains largely preserved, with GenEval scores decreasing by $0.04$ for FUDOKI and $0.03$ for UniDFlow-4B. These results suggest that our framework improves visual fidelity and aesthetics while largely retaining discrete compositional priors.

\FUDOKIFigure

\subsection{Multimodal Understanding}

Table~\ref{tab:understanding_results} evaluates whether adapting a discrete flow-matching (DFM) unified model to a continuous flow-matching (CFM) space compromises pretrained semantic understanding. The DFM$\rightarrow$CFM transition introduces only minor performance changes. For UniDFlow-4B, performance decreases by $0.28$, $0.24$, $0.30$, $0.70$, and $0.40$ percentage points on MMBench, MMMU, MM-Vet, MathVista, and MMVP, respectively. Averaging the relative changes across these five benchmarks yields approximately 0.46\% degradation, corresponding to 99.54\% retention of the original performance, with relative degradation below $1\%$ on each benchmark. FUDOKI similarly exhibits approximately 0.27\% average relative degradation, corresponding to 99.73\% retention based on the displayed scores, while improving on MMVP from $67.72$ to 68.01. Overall, results indicate that the representation transition modifies visual generation while largely preserving pretrained multimodal understanding and performance on reasoning tasks.

\section{Qualitative Results}
\label{sec:qualitative}

Figures~\ref{fig:UniDFlowFigure} and~~\ref{fig:FUDOKIFigure}  compare the original discrete UniDFlow-4B and FUDOKI models with their continuous counterparts obtained using our \ours{} respectively. Across diverse prompts and complex scenes, the continuous variants consistently exhibit improved local fidelity while preserving the semantic composition of the corresponding discrete models. For FUDOKI (food example), the continuous variants better resolve the croissant's flaky layers. In the UniDFlow portrait example, continuous generation produces more refined facial details, hair strands, and accessories. Similar improvements are visible for the hummingbird, where feather structure, specular coloration, and flower details become sharper and more coherent. Importantly, such improvements are observed for both FUDOKI and UniDFlow-4B rather than being specific to a single backbone.

\TableAblation

\UndResultsFigure

Additionally, Figure~\ref{fig:UndResults} qualitatively confirms that the DFM$\rightarrow$CFM transition using \ours{} preserves the model's multimodal understanding, retaining strong performance across diverse reasoning and counting tasks.

\section{Ablations}
\label{sec:ablation}
Table~\ref{tab:ablation_study} evaluates the key design choices of \ours{}: (i) the necessity of distillation beyond standard flow matching; (ii) sensitivity to teacher choice and capacity; and (iii) the role of parameter-efficient attention adaptation in accommodating representation shift while preserving pretrained backbone semantics. By default, we utilize UniDFlow-4B as the discrete unified backbone, Qwen Image as the continuous teacher, and LoRA adapters for adaptation.

\noindent \textbf{Impact of velocity distillation.}
Removing the continuous teacher to train the student solely against the rectified-flow target $\boldsymbol{\epsilon}-\mathbf{z}_0$ (using $\mathcal{L}_{\mathrm{FM}}$ without the velocity-distillation term $\mathcal{L}_{\mathrm{VD}}$) consistently degrades all generation metrics: GenEval drops from $0.92$ to $0.89$, COCO-30K FID increases from $8.92$ to $10.69$, CLIPScore decreases from $38.23$ to $36.31$, and aesthetic score falls from $6.21$ to $5.58$. These results validate DOVD: while both terms supervise velocity, $\mathcal{L}_{\mathrm{FM}}$ regresses the conditional velocity $\epsilon - z_0$, forcing the student to recover the marginal field from a high-variance signal where noise cancels only in expectation. Conversely, $\mathcal{L}_{\mathrm{VD}}$ directly transfers continuous rendering behavior by matching a learned, low-variance field rather than re-estimating it from data. Consequently, isolating $\mathcal{L}_{\mathrm{FM}}$ reduces perceptual quality (Table~\ref{tab:ablation_study}).

\noindent \textbf{Impact of the teacher models.}
Replacing the default teacher with Flux-1.0 and Flux-2 variants yields nearly unchanged semantic performance: GenEval varies only between $0.91$--$0.92$, MMBench between $90.23$--$90.58$, and MMMU between $74.17$--$74.67$. Conversely, generation quality is more teacher-sensitive, with Flux-2 32B achieving the best COCO-30K FID ($8.33$) and aesthetic score ($6.24$). This confirms our hypothesis: the teacher primarily transfers continuous rendering dynamics, while the frozen unified backbone preserves semantic grounding. Because no single teacher is uniformly optimal, we adopt Qwen Image to provide the best overall balance between semantic retention and perceptual quality.

\noindent \textbf{Full fine-tuning vs. LoRA.}
We examine whether the representation transition requires full-backbone updates. Full fine-tuning yields only marginal generation gains (COCO-30K FID $8.92\rightarrow8.76$, CLIPScore $38.23\rightarrow38.52$), while substantially degrading semantic understanding, with MMBench dropping from $90.92$ to $86.01$ and MMMU from $74.26$ to $69.28$. Similar degradation without teacher distillation indicates semantic drift from unrestricted updates. Thus, LoRA provides sufficient flexibility for the discrete-to-continuous representation shift while better preserving pretrained semantic knowledge.

\section{Conclusion}
\label{sec:conclusion}
We introduce a framework (\ours{}) to transform pretrained discrete unified multimodal models into continuous rectified-flow generators without relearning established semantic knowledge. By freezing the backbone and adapting only lightweight continuous interfaces and low-rank attention modules, our approach transfers high-fidelity rendering dynamics from a continuous teacher via Dual-Objective Velocity Vector Distillation. Evaluations confirm \ours{} yields improvements in visual distribution similarity and aesthetics with negligible degradation in precise compositional control. Ultimately, this establishes a scalable pathway for future unified architectures to achieve superior continuous visual synthesis while retaining the rigorous logical grounding of discrete language modeling.

\section*{Acknowledgments}
This research is based upon work supported by the NSF CAREER Award \#2542328 and the U.S. Defense Advanced Research Projects Agency (DARPA) under award HR001125C0303. The views and conclusions contained herein are those of the authors and should not be interpreted as necessarily representing the official policies, either expressed or implied, of NSF, DARPA, or the U.S. Government. The U.S. Government is authorized to reproduce and distribute reprints for governmental purposes notwithstanding any copyright annotation therein. 

{
\small
\bibliographystyle{ieeenat_fullname}
\bibliography{main}
}

\appendix

\maketitlesupplementary

\SuppFigure

\section{Experimental Details} \label{sec:supp_experimental_sec}

\subsection{Training Data}

For training, we construct a large-scale image--text corpus from three datasets: LAION-20M Recap~\cite{li2024laion}, OpenGPT-4o-Image~\cite{chen2025opengpt}, and DRAGON~\cite{bertazzini2025dragon}, comprising approximately 6M image-text pairs in total. LAION-20M Recap and OpenGPT-4o-Image provide high-quality images paired with detailed, information-rich captions, offering strong supervision for learning fine-grained text-image correspondence. We further incorporate DRAGON, a synthetically curated dataset generated using text-to-image models, to increase the diversity of visual concepts, compositions, and generation patterns represented during training.

\begin{table*}[t]
    \centering
    \caption{\textbf{Summary of evaluation benchmarks.}
    We report the number of evaluation examples, input--output format,
    task category, and evaluation metric. For image-generation
    benchmarks, each example denotes a text prompt, while for
    multimodal-understanding benchmarks, each example corresponds to
    a visual question-answering instance. Arrows indicate whether
    higher ($\uparrow$) or lower ($\downarrow$) values are preferred.\looseness-1}
    \label{tab:eval_datasets}
    \vspace{-0.3cm}
    \resizebox{\linewidth}{!}{
    \begin{tabular}{lcccc}
        \toprule
        \textbf{Benchmark} &
        \textbf{\# Examples} &
        \textbf{Format} &
        \textbf{Task} &
        \textbf{Evaluation Metric} \\
        \midrule

        \multicolumn{5}{c}{\textit{Image Generation}} \\
        \midrule

        GenEval~\cite{ghosh2023geneval}
        & 553
        & Text Prompt
        & Compositional T2I
        & GenEval Acc. $\uparrow$ \\

        COCO-30K~\cite{lin2014microsoft}
        & 30,000
        & Image Caption
        & Text-to-Image
        & FID $\downarrow$, DINOv2-FID $\downarrow$ \\

        Aesthetic-Eval~\cite{zhang2025diffusion}
        & 2,781
        & Image--Text Pair
        & High-Resolution T2I
        & FID $\downarrow$, CLIPScore $\uparrow$,
          Aesthetic Score $\uparrow$ \\

        \midrule
        \multicolumn{5}{c}{\textit{Multimodal Understanding}} \\
        \midrule

        MMBench v1.1~\cite{liu2024mmbench}
        & 3,217
        & Multiple Choice
        & General Multimodal QA
        & CircularEval Acc. $\uparrow$ \\

        MMMU~\cite{yue2024mmmu}
        & 11,550
        & MCQ + Open-ended
        & Expert-level Reasoning
        & Overall Acc. $\uparrow$ \\

        MM-Vet~\cite{yu2023mm}
        & 218
        & Open-ended
        & Integrated Multimodal QA
        & GPT-4 Score $\uparrow$ \\

        MathVista~\cite{lu2024mathvista}
        & 6,141
        & MCQ + Open-ended
        & Visual Mathematical Reasoning
        & Accuracy $\uparrow$ \\

        MMVP~\cite{tong2024eyes}
        & 300
        & Multiple Choice
        & Fine-grained Visual Perception
        & Pair Accuracy $\uparrow$ \\

        \bottomrule
    \end{tabular}}
\end{table*}

\subsection{Implementation Details}

We train FlowShift on 8 NVIDIA H100 GPUs with 80 GB memory each. Training uses a per-GPU batch size of 32 with 4 gradient-accumulation steps, resulting in an effective batch size of 1,024. We optimize the model for 5K steps using the Muon~\cite{liu2025muon} optimizer with an initial learning rate of ($1\times10^{-5}$). The learning rate is linearly warmed up for the first 2K steps and subsequently decayed using a cosine annealing schedule. For an efficient parameter adaptation, we employ LoRA with rank (r=32) and scaling factor ($\alpha = 16$). Image-generation captions are truncated to a maximum sequence length of 500 tokens.

During training, we preserve the native aspect ratios of images rather than resizing all samples to a fixed resolution. This exposes the model to diverse spatial configurations and improves the generalization of its rotary positional embeddings (RoPE) across different resolutions and aspect ratios. For sampling, we use an Euler discrete scheduler with a classifier-free guidance (CFG) scale of (8.0), a true-CFG scale of (4.0), and negative-prompt guidance with a scale of (4.0). Both training and inference are implemented using the Hugging Face \texttt{diffusers} framework.

\subsection{Evaluation Datasets and Metrics}

We evaluate our model across both image generation and multimodal
understanding, as summarized in Tab.~\ref{tab:eval_datasets}.
For image generation, we consider three complementary benchmarks.
GenEval~\cite{ghosh2023geneval} consists of 553 compositional prompts
designed to evaluate object presence, counting, color attribution,
spatial relationships, and attribute binding. COCO-30K~\cite{lin2014microsoft}
contains 30,000 caption-based generation prompts sampled from MS-COCO
and is used to measure distribution-level image quality using
FID~\cite{heusel2017gans} and DINOv2-FID. We additionally evaluate on
Aesthetic-Eval~\cite{zhang2025diffusion}, comprising 2,781 high-resolution
image--text pairs, to assess perceptual quality, text--image alignment,
and aesthetic fidelity using FID, CLIPScore~\cite{hessel2021clipscore},
and Aesthetic Score~\cite{zhang2025diffusion}.

For multimodal understanding, we evaluate on MMBench
v1.1~\cite{liu2024mmbench}, which contains 3,217 multiple-choice
visual reasoning questions, and MMMU~\cite{yue2024mmmu}, comprising
11,550 expert-level multimodal questions spanning both multiple-choice
and open-ended formats. We further evaluate on MM-Vet~\cite{yu2023mm},
which contains 218 open-ended questions covering integrated
vision-language capabilities, and MathVista~\cite{lu2024mathvista},
which contains 6,141 multiple-choice and free-form questions requiring
visual mathematical reasoning. Finally, MMVP~\cite{tong2024eyes}
contains 300 multiple-choice questions constructed from 150 visually
challenging image pairs and evaluates fine-grained visual perception.
Together, these benchmarks measure whether continuous adaptation
preserves the semantic perception and reasoning capabilities of the
unified backbone while improving its image-generation ability.

\end{document}